\documentclass[journal]{IEEEtran}

\usepackage{graphicx}
\usepackage{amssymb}
\usepackage{amsmath}

\usepackage{algorithm}
\usepackage{algorithmic}
\usepackage{enumerate}
\usepackage{threeparttable}
\usepackage{multirow}
\usepackage[colorlinks]{hyperref}

\begin{document}

\author{Zhifei Li,
	Feng Qiu,
        Yiran Wang,
        Yujing Xia,
        Kui Xiao,
        Miao Zhang,
        and Yan Zhang

\IEEEcompsocitemizethanks{
\IEEEcompsocthanksitem This work was supported in part by the National Natural Science Foundation of China (No. 62207011, 62377009, 62407013), the Natural Science Foundation of Hubei Province of China (No. 2025AFB653). (Corresponding authors: Miao Zhang, Yan Zhang)
\IEEEcompsocthanksitem Z. Li, F. Qiu, Y. Wang, Y. Xia, M. Zhang, K. Xiao, and Y. Zhang are with the School of Computer Science, Hubei University, Wuhan 430062, China, Hubei Key Laboratory of Big Data Intelligent Analysis and Application (Hubei University), Wuhan 430062, China, and also with the Key Laboratory of Intelligent Sensing System and Security (Hubei University), Ministry of Education, Wuhan 430062, China. 
	}

} 

\title{Integrating Object Interaction Self-Attention and GAN-Based Debiasing for Visual Question Answering}

\maketitle

\markboth{IEEE Transactions on Multimedia}
{Li \MakeLowercase{\textit{Li et al.}}: IOG-VQA}

\begin{abstract}

Visual Question Answering (VQA) presents a unique challenge by requiring models to understand and reason about visual content to answer questions accurately. Existing VQA models often struggle with biases introduced by the training data, leading to over-reliance on superficial patterns and inadequate generalization to diverse questions and images. This paper presents a novel model, IOG-VQA, which integrates Object Interaction Self-Attention and GAN-Based Debiasing to enhance VQA model performance. The self-attention mechanism allows our model to capture complex interactions between objects within an image, providing a more comprehensive understanding of the visual context. Meanwhile, the GAN-based debiasing framework generates unbiased data distributions, helping the model to learn more robust and generalizable features. By leveraging these two components, IOG-VQA effectively combines visual and textual information to address the inherent biases in VQA datasets. Extensive experiments on the VQA-CP v1 and VQA-CP v2 datasets demonstrate that our model shows excellent performance compared with the existing methods, particularly in handling biased and imbalanced data distributions highlighting the importance of addressing both object interactions and dataset biases in advancing VQA tasks. Our code is available at \url{https://github.com/HubuKG/IOG-VQA}.
\end{abstract}

\begin{IEEEkeywords}
Visual Question Answering,
Self-attention Mechanism,
GAN-based Debiasing
\end{IEEEkeywords}

\section{Introduction}
\label{section1}

\begin{figure}[t]
    \centering
    \includegraphics[width=0.8\linewidth]{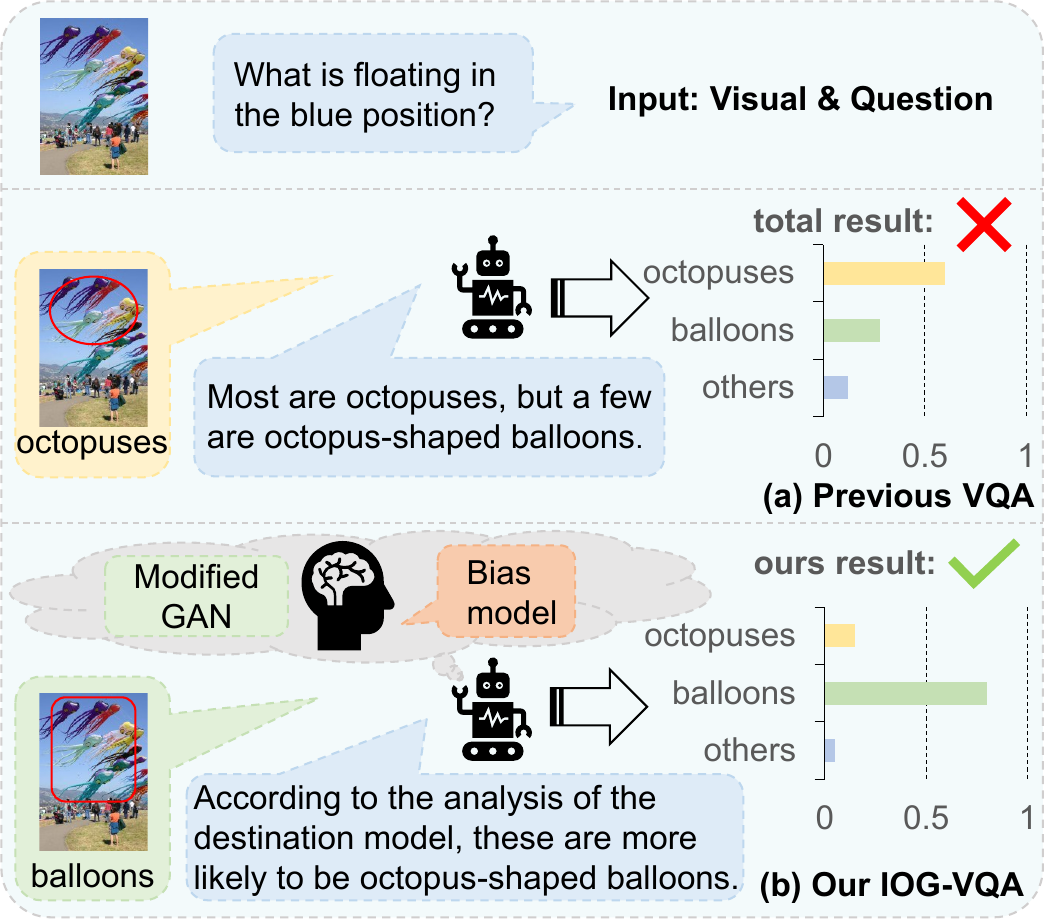}
    \caption{Comparison between Previous VQA Models and Our IOG-VQA Model.  Traditional VQA models often misinterpret questions due to training data biases. Our IOG-VQA model addresses this by using an object interaction self-attention module and a Modified GAN for more accurate and contextually appropriate answers.}
    \label{fig:motivation}
\end{figure}

\IEEEPARstart{V}{isual} Question Answering (VQA) \cite{1} is an interdisciplinary field that combines the challenges of computer vision and natural language processing to generate accurate answers to questions about images. This task requires a deep understanding of both the visual content and the contextual nuances posed by the questions, making it a complex and demanding research area. Despite significant advancements in recent years, current VQA models often struggle with biases introduced by training data \cite{2,3,4}, leading to an over-reliance on superficial patterns and correlations rather than genuine visual reasoning and understanding. For example, consider the question “What is floating in the blue position?”  In Fig. \ref{fig:motivation}, the model may incorrectly predict octopuses because the training data often associates blue locations with the ocean and octopuses, relying on surface patterns instead of truly reasoning about the visual and textual information. Recent works \cite{5,6,7,8,9} have proposed many constructive debias methods based on ensemble models for such problems.

In these ensemble model debiasing methods, they will use in addition to the model itself, the addition of additional models to help their method to achieve the detection of bias and reduce the additional impact of bias on the model itself. For example, some popular solutions such as \cite{7,9} are to use additional pure question-answer branches to synthesize the language given before training, in this way to generate as much bias as possible, and then remove these pure language branches during the testing process to reduce the bias. For example, in \cite{6,8}, the question-answering Model is used to train a perfect VQA Model, and this bias model mainly senses and expresses the bias through the language before teaching the VQA Model to reduce the influence of bias. Many existing debias methods based on ensemble models such as \cite{6,7,8,10} perform auxiliary training and make predictions from question or image unimodality respectively, or rely on precomputed label statistics, such as GGE-D \cite{8} and LMH \cite{7}, for training.

However, solely relying on question-answer pairs or precomputed labels has limitations. The training data and the derived biases are finite and cannot fully encapsulate the diverse and intricate biases that may influence the VQA Model. A fundamental challenge for many VQA Models is their sensitivity to dataset biases \cite{11}, which can significantly affect their performance and generalization capabilities. These biases often stem from imbalanced training data, where certain answer types or visual contexts are overrepresented. Consequently, models tend to memorize these frequent patterns rather than learning to reason based on the actual content of the images and questions. This overfitting to dataset biases limits the applicability of VQA models in real-world scenarios, where diverse and unbiased reasoning is essential.

To solve the above problems, we propose a novel model named IOG-VQA (Integrating Object Interaction Self-Attention and GAN-Based Debiasing for Visual Question Answering). Our model introduces an object interaction self-attention module to capture complex inter-object relationships, leading to a holistic understanding of the visual scene for more accurate answers. For question processing, we use self-attention mechanisms alongside alphabet and word embeddings to better capture the nuances of the queries. To reduce dataset bias, we integrate a Modified Generative Adversarial Network (based GAN \cite{12}) combined with Feature Transformer Networks, which generates unbiased data distributions through realistic perturbations. This helps the model learn robust features less dependent on biased patterns. By back training to distinguish between biased and unbiased samples, we enhance generalization. We also use knowledge distillation \cite{13} to transfer debiased knowledge, reducing overall bias. A dedicated debiasing loss function is introduced to ensure accurate and unbiased predictions. We validate our model's performance through extensive experiments on two common VQA datasets without external annotations or additional processing, demonstrating the superiority of our method.

Our contributions can be summarized as follows:

\begin{itemize}
    \item We propose an object interaction self-attention mechanism to effectively model the interactions between objects in images, enabling the model to develop a more comprehensive and nuanced understanding of the visual scene, which in turn enhances its ability to generate accurate and contextually appropriate answers.
    \item We introduce a novel debiasing method, IOG-VQA, based on the development of the debiasing framework of Modified GAN and use the modified GAN to generate unbiased data distributions, significantly improving the ability of the model to handle imbalanced and biased datasets.
    \item We conduct extensive experiments on VQA-CP v1 and VQA-CP v2 datasets to validate the proposed method with UpDn as the baseline architecture. The experimental results show that IOG-VQA shows a relatively excellent level, especially in the scenes with biased and imbalanced data.
\end{itemize}

The remainder of this paper is organized as follows: Section \ref{section2} provides a detailed review of related literature. Section \ref{section3} elaborates on the specifics of our proposed model. Section \ref{section4} presents the experimental results and analysis in detail. Finally, Section \ref{section5} summarizes the paper and offers concluding remarks.

\section{Related work}
\label{section2}


Visual Question Answering (VQA) \cite{1,15,16} combines computer vision and natural language processing to develop models that can answer questions about images. The complexity of understanding the visual content and contextual nuances of the problem has led to a variety of research efforts, and there is currently a lot of good work \cite{17,18,19,20,21} that makes the performance of its methods humanlike, with this paper focusing specifically on debias strategies, attention mechanisms, and object interactions, generating models.

\subsection{Bias Reduction Techniques in VQA}

A major challenge in VQA is addressing training dataset biases. Agrawal et al. \cite{11} introduced the VQA-CP dataset to expose models to different answer distributions during training and testing, promoting robust learning. Some studies \cite{22,23,24} generate counterfactual data by hiding or transforming key images and words to produce ground truth answers. Other approaches \cite{25,26,27} introduce negative samples without any annotations to reduce biases. Additionally, Data Augmentation (DA) methods, such as KDDAug \cite{28}, DDG \cite{DDG}, and Mutant \cite{23}, indeed offer valuable strategies for enhancing model robustness by generating extra training samples beyond the original training expanding training data diversity. Methods such as KDDAug \cite{28,29} replacing images or questions and Mutant \cite{23} using external correction networks to expand datasets and alter distributions have been employed, DDG \cite{DDG} propose a method to dig out discrimination information from generated samples. While these approaches demonstrate effectiveness, CF-VQA \cite{10}, Genb \cite{33} emphasizes that modifying the training distribution (e.g., via synthetic data generation) could deviate from the foundational purpose of the VQA-CP benchmark. The benchmark specifically evaluates model robustness to biases under unaltered training conditions. To maintain alignment with this principle, our experimental comparisons focus on methods adhering to the original dataset design, rather than those involving training distribution shifts.

For ensemble-based methods, such as LMH \cite{7} utilize counterfactual data augmentation to create modified versions of questions and images, reducing model reliance on dataset biases. The GGE \cite{8} framework uses models identical to the target model to learn biases, while Cadene et al. developed RUBi \cite{6}, which incorporates a bias reduction module to mitigate biases learned from question-only training. Another method, AREG \cite{9}, uses separate question-only models to capture biases and further debias the main model. Niu et al. \cite{10} introduced a counterfactual inference framework to mitigate biases, and later \cite{30} improved both out-of-distribution generalizability and performance through introspective knowledge distillation. KV and Mittal \cite{31} employed a visual question encoder to embed visual information into question representations. The Multi-modal Debiasing Model (MDDC) \cite{32} proposed a strategy to subtract bias scores and introduced two bias learning branches to detect and address bias issues. The GenB \cite{33} framework employs generative models to create debiased training samples, highlighting the potential of generative techniques. TDS \cite{TDS} dynamically mitigate bias during inference by constructing negative samples and entropy filtering, enabling OOD adaptation without altering training processes. PW-VQA \cite{PW-VQA} addresses vision-language bias via causal graphs and counterfactual inference, reducing cross-modality confounding effects while enhancing numerical answer accuracy. CVIV \cite{50} proposes an unbiased VQA model that leverages instrumental variables to enhance the causal effect between visual features and answers, aiming to improve the model's ability to focus on relevant visual information and reduce reliance on spurious correlations.

In contrast to these approaches, our proposed IOG‐VQA distinguishes itself by integrating an object interaction self‐attention module with a GAN‐based debiasing framework. Rather than relying solely on counterfactual data augmentation or ensemble methods to mitigate bias, IOG‐VQA directly learns unbiased data distributions through adversarial training while simultaneously capturing complex inter‐object relationships. This dual mechanism not only avoids altering the original data distribution but also enhances generalization across diverse visual and linguistic contexts, thereby offering a more balanced and comprehensive solution to bias reduction in VQA tasks.

\subsection{Attention Mechanisms and Object Interaction in Visual Question Answering}
Yang et al. introduced SANs \cite{34}, which apply multiple attention layers to refine attention to image regions gradually. Anderson et al. \cite{14} enhanced this method by proposing bottom-up and top-down attention models, combining object detection features with attention mechanisms to improve visual grounding and problem understanding. Among vision-based methods \cite{35,36} utilize human annotations as supervision to enhance the accuracy of attention maps \cite{37,38}, there are also methods to improve visual attention by annotating additional annotations \cite{39,40}. Jin et al. \cite{41} introduced a multi-interaction network with object relations, enhancing video question answering by combining multi-modal and multi-level interactions. A-ATT \cite{A-ATT} introduces multi-round accumulated attention that progressively integrates cross-modal cues through self-attention guidance and noised training, reducing visual-textual redundancies for robust grounding. PFOS \cite{PFOS} introduces dense cross-attention between visual grids and textual words through bidirectional guidance modules (LGV/VGL), enabling proposal-free joint reasoning for expression comprehension and generation. S3C \cite{S3C} employs self-critical attention to align answer-explanations via answer-rewarded reinforcement, integrating semi-supervised learning to enhance logical consistency without full annotation dependency.

In contrast to these approaches, our proposed IOG-VQA model not only utilizes attention mechanisms to refine visual representations but also incorporates an object interaction self-attention module that explicitly models inter-object relationships. While methods like SANs, bottom-up/top-down attention, A-ATT, PFOS, and S3C primarily focus on enhancing visual grounding and integrating cross-modal cues, our approach uniquely combines these attention strategies with a GAN-based debiasing framework. This integration allows IOG-VQA to generate unbiased data distributions while capturing complex object interactions, thereby addressing both visual reasoning and bias reduction in a unified manner. This dual mechanism clearly differentiates our method from the existing works and underscores its novelty and contribution to the field of VQA.

\subsection{Generative Adversarial Networks (GANs) in VQA}
Generative Adversarial Networks (GANs) \cite{12} have shown promise in improving VQA by generating synthetic data to help reduce bias and model generalization. Reed et al. introduce a text-to-image synthesis model based on generative adversarial networks \cite{42}, capable of generating realistic images from detailed text descriptions. Furthermore, Chen et al. \cite{17} use GANs to create adversarial examples, enhancing VQA models’ robustness against misleading correlations and significantly improving performance on standard benchmarks. The Transformer architecture has also been adapted for the VQA task \cite{21,43}, showing significant performance gains by leveraging self-attention to model global dependencies in visual and textual.

In contrast to the aforementioned approaches that primarily focus on generating adversarial examples or synthetic images, our proposed IOG-VQA method employs a modified GAN framework that is tightly integrated with an object interaction self-attention module. This dual mechanism not only generates unbiased data distributions to mitigate dataset biases but also captures complex inter-object relationships. Consequently, our approach offers a more balanced fusion of visual and textual cues, thereby enhancing the overall generalization and robustness of the VQA model.

The VQA field has advanced significantly through the incorporation of attention mechanisms, bias reduction strategies, and generative models. Despite these advances, challenges remain in achieving unbiased and generalizable VQA systems. Our proposed IOG-VQA model integrates object interaction self-attention with GAN-based debiasing, showing promising results on benchmark datasets. Future research will continue exploring the synergy between these techniques to create more robust and interpretable VQA models.

\begin{figure*}
    \centering
    \includegraphics[width=0.8\linewidth]{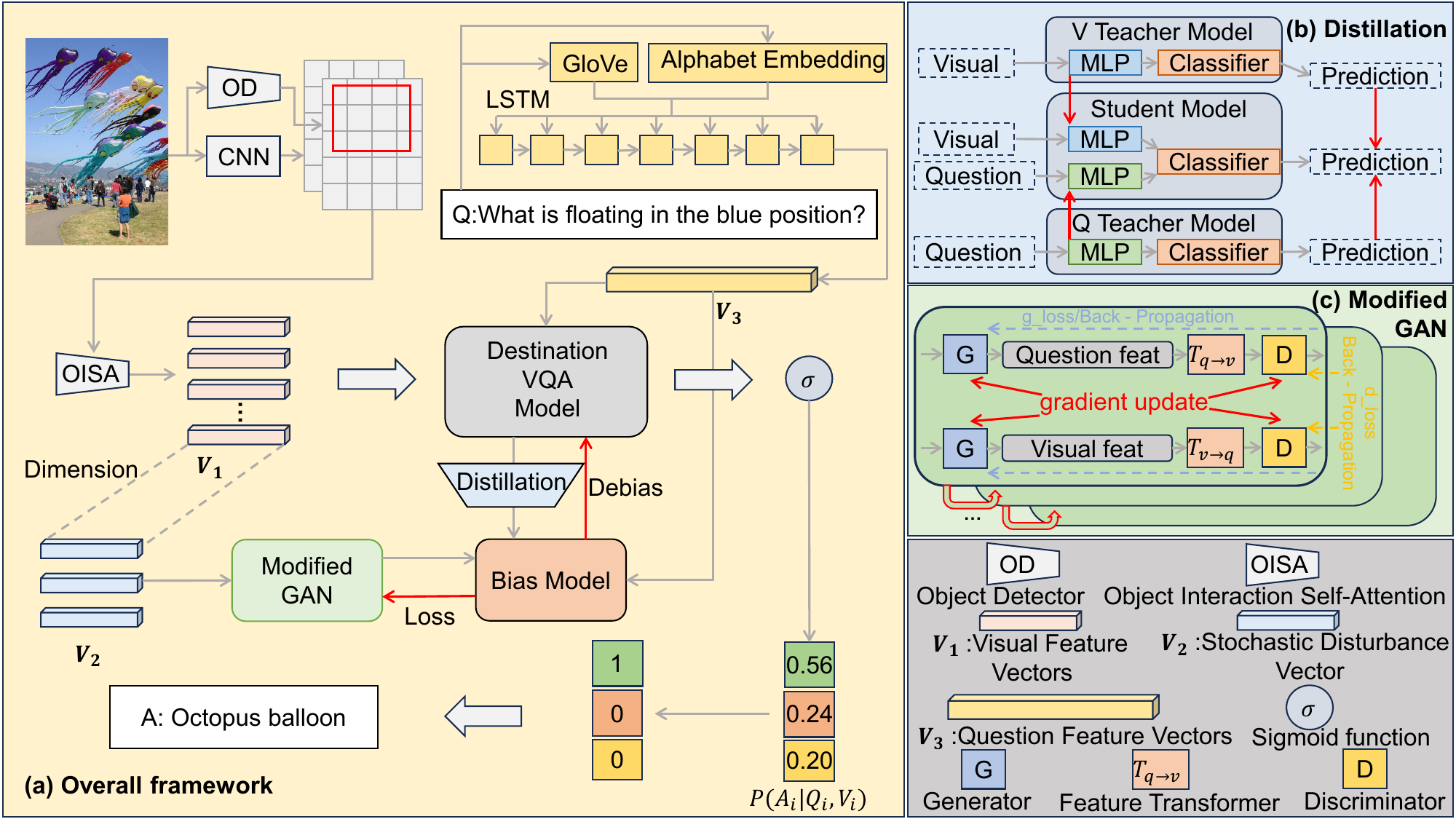}
    \caption{(a) shows the overview of our proposed IOG-VQA method with training examples related to ‘What is floating in the blue position?’. (b) shows the framework of the knowledge distillation approach. (c) shows the framework of the Modified GAN approach. The bottom right corner is the explanation of the legend.}
    \label{fig:overview}
\end{figure*}

\section{Proposed Method}
\label{section3}


  In this section, we will describe our approach IOG-VQA in detail, it includes question processing, visual processing, bias model training, model prediction, and probabilistic results. In Fig. \ref{fig:overview}, an overview of our model IOG-VQA is presented.

\subsection{Question processing}
We use alphabet embeddings $\mathbf{Q}^{a} \in \mathbb{R}^{i \times a \times d_{a}}$ and word embeddings $\mathbf{Q}^{w} \in \mathbb{R}^{i \times d_{w}}$ to represent a question $\mathbf{Q}$ with $i$ words, where $d_{a}$ and $d_{w}$ denote the dimensions of the alphabet embeddings and word embeddings, respectively. The word embedding function is initialized with pre-trained GloVe \cite{44}, while the alphabet embedding function is randomly initialized. Given the alphabet embeddings and word embeddings, we represent the question embedding using a self-attention mechanism $s(\cdot, \cdot)$. The alphabet embeddings are further processed by a function $c(\cdot)$, which consists of a 2D convolutional layer. Then, we feed the question embedding $\mathbf{Q}$ into LSTM function $l(\cdot)$, obtaining the question feature vectors $\mathbf{Q}^{v}$,
\begin{equation}
    \mathbf{Q}^{v} = l \left( s \left( \mathbf{Q}^{w}, c \left( \mathbf{Q}^{a} \right) \right) \right).
\end{equation}

\subsection{Visual processing}


Given a video frame, we first extract its feature map $I$ using EfficientNet \cite{45}. Next, for this frame, the YOLO detector is employed to identify objects and generate $n$ bounding boxes. After that, RoIAlign \cite{46} is applied to extract refined object features $\mathbf{O}_i$ from the proposal regions, expressed in the following equation:

\begin{equation}
    \mathbf{O}_i = \text{RoIAlign}(I, b_i), \quad i = 1,2, \ldots, n,
\end{equation}
where $b_i$ represents the bounding box of the $i$-th object.
Then, these features are further processed through a fully connected layer with an ELU activation function to reduce their dimensionality.

Each object feature is self-attention calculated to achieve information transfer between objects:
\begin{equation}
    \widehat{\mathbf{O}}_i = \sum_{j=1}^{n} {\frac{\exp({\frac{(\mathbf{W}_q \mathbf{O}_i)^\top (\mathbf{W}_k \mathbf{O}_j)}{\sqrt{d}}})}{\sum_{k=1}^{n} \exp(e_{ik})}}(\mathbf{W}_v \mathbf{O}_j),
\end{equation}

here, $\mathbf{W}_q$, $\mathbf{W}_k$, and $\mathbf{W}_v$ are the weight matrices of the query, key, and value, respectively, and $d$ is the scaling factor (the dimension of the key). The object feature after object interaction from the attention module is called region feature $\mathbf{F}_R$.

Global average pooling of frame features to generate global feature $\mathbf{F}_G$. The use of self-attention mechanisms to fuse global and regional features:
First, the global feature $\mathbf{F}_G$ is copied $n$ times and joined with the regional feature $\mathbf{F}_R$:
\begin{equation}
\widetilde{\mathbf{F}}_R = [\mathbf{F}_R; \mathbf{F}_G \otimes \mathbf{1}_n].
\end{equation}

Then self-attention calculation is performed on the spliced features:
\begin{equation}
\widehat{\mathbf{F}}_R = \text{softmax}\left(\frac{\widetilde{\mathbf{F}}_R \widetilde{\mathbf{F}}_R^\top}{\sqrt{d}}\right) \mathbf{F}_R.
\end{equation}

Finally, features after attention are fused using 1D convolution and ELU activation function, producing the visual features ${\mathbf{F}}_V$.

\subsection{Bias Model Training}

In visual question answering (VQA) tasks, Weighted Cross-Entropy Loss is mainly used to deal with class imbalance problems. Specifically, some categories of answers may be more common than others, which can cause the model to tend to predict these common categories and ignore predictions for rare categories. By assigning different weights to different classes, the weighted cross-entropy loss allows the rare class to receive more attention in the loss calculation, thereby helping the model reduce bias and improve the prediction ability for all classes:
\begin{equation}
L_{\text{WCE}}(y, \hat{y}) = -\sum_{i=1}^{N} w_i \left[ y_i \log(\hat{y}_i) + (1 - y_i) \log(1 - \hat{y}_i) \right],
\end{equation}
where $y_i$ is the true label of the $i$-th sample, $\hat{y}_i$ is the prediction probability of the $i$-th sample, $w_i$ is the weight of the $i$-th sample, and $N$ is the total number of samples.

Visual features $\mathbf{V}_1$ and question features $\mathbf{V}_3$ are input into the Destination VQA Model to generate preliminary predictions. The model is then debiased using a distillation technique, which approximates the output distribution of larger, complex models (teacher models) by training a smaller model (student model). In this paper, the distillation process involves the visual teacher model (V Teacher Model), the student model, and the question teacher model (Q Teacher Model). The V Teacher Model and Q Teacher Model handle visual and question features, respectively, and guide the student model’s training to integrate visual and textual information, reduce bias, and improve overall performance. The process involves: Training the V Teacher Model and Q Teacher Model to accurately predict visual and question features, producing outputs $P_{t}^{(v)}$ and $P_{t}^{(q)}$. Using the Student Model to process these features and obtain its output $P_{s}$. Calculating the distillation loss to guide the student model in learning from the teacher models. The corresponding formulas are as follows:

\begin{equation}
    \mathcal{L}_{distill}=\alpha\cdot KL(P_t^{(v)}||P_s)+\beta\cdot KL(P_t^{(q)}||P_s),
\end{equation}
\begin{equation}
    KL(P||Q)=\sum_iP(i)\log\left(\frac{P(i)}{Q(i)}\right),
\end{equation}
here, $\text{KL}(P \| Q)$ stands for Kullback-Leibler Divergence \cite{47}, which measures the difference between two probability distributions. $\alpha$ and $\beta$ are hyperparameters that regulate the weight of the visual teacher model loss and the question teacher model loss. $P_i$ and $Q_i$ represent class $i$ probabilities respectively.

In this paper, a modified generative adversarial network (Modified GAN) is used to further reduce bias and improve model performance. Specifically, this paper's Modified GAN process incorporates Feature Transformer Networks to better handle visual and problem features. The following is the detailed process and the corresponding formula and explanation:
The generator generates stochastic disturbance features $\mathbf{V}_2$ to make these disturbance features realistic enough to fool the discriminator. Feature Transformer Networks are used to convert question features into visual features or visual features into question features. The purpose is to better fuse and process these two types of features so that the generator can generate more realistic features:
\begin{equation}
\mathbf{V}_3' = T_{q \rightarrow v}(\mathbf{V}_3),\mathbf{V}_1' = T_{v \rightarrow q}(\mathbf{V}_1),
\end{equation}
where $\mathbf{V}_3'$ is the transformed visual feature, $\mathbf{V}_1'$ is the transformed question feature, $T_{q \rightarrow v}$ and $T_{v \rightarrow q}$ are the feature transformer networks. The discriminator discriminates between the true and generated features to obtain $D(\mathbf{V}_1)$ and $D(\mathbf{V}_2)$. 
The modified GAN can be represented as follows:
The generator tries to minimize the discriminant probability of the generated feature to improve the authenticity of the generated feature:
\begin{equation}
\mathcal{L}_{G} = -\mathbb{E}[\log D(\mathbf{V}_2)].
\end{equation}

\begin{table}[t]
\centering
\caption{The amount of data on the VQA-CP v1 and VQA-CP v2 train and test sets.}
\label{table:1}
\begin{tabular}{cccc}
\hline
\textbf{Dataset} & \textbf{images} & \textbf{questions} & \textbf{answers} \\
\hline
VQA-CP v1 train & 118K & 245K & 2.5M \\
VQA-CP v1 test & 87K & 125K & 1.3M \\
\hline
VQA-CP v2 train & 121K & 438K & 4.4M \\
VQA-CP v2 test & 98K & 220K & 2.2M \\
\hline
\end{tabular}
\end{table}

The loss of the feature transformer networks is used to minimize the difference between transformer feature and real feature, ensuring that the feature generated by the feature transformer networks is as close as possible to the real feature, the formula for $\mathcal{L}_{T_{v \rightarrow q}}$ is similar to $\mathcal{L}_{T_{q \rightarrow v}}$ as follows:
\begin{equation}
\mathcal{L}_{T_{q \rightarrow v}} = \mathbb{E}\left[ \| \mathbf{V}_1 - T_{q \rightarrow v}(\mathbf{V}_3) \|^2 \right],
\end{equation}
\begin{align}
\mathcal{L}_{T_{v \rightarrow q}} = \mathbb{E}\left[ \| \textbf{V}_3 - T_{v \rightarrow q}(\textbf{V}_1) \|^2 \right].
\end{align}

The discriminator tries to maximize the discriminant probability for the real feature and minimize the discriminant probability for the generated feature:
\begin{equation}
\mathcal{L}_{D} = -\mathbb{E}[\log D(\mathbf{V}_1)] - \mathbb{E}[\log(1 - D(\mathbf{V}_2))],
\end{equation}
the discriminator loss ($\mathcal{L}_D$) and the generator loss ($\mathcal{L}_G$) are backpropagated to update the discriminator and the generator, respectively. After applying gradient clipping, both networks are updated via gradient descent, and these updated models are subsequently used in the next training iteration—ensuring continuous improvement in distinguishing real from generated features and in generating more realistic, unbiased features over time. 
Through adversarial training, the discrimination ability of the discriminator is gradually enhanced, and the generation ability of the generator is also continuously improved. 

The GAN total loss function combines discriminator loss, feature transformer loss, and generator loss, and realizes the cooperative optimization of each part by adjusting the weight of different loss terms and finally improves the performance of the model:
\begin{equation}
\mathcal{L}_{\text{GAN}} = \mathcal{L}_{D} + \mathcal{L}_{G} + \lambda_1 \mathcal{L}_{T_{q \rightarrow v}} + \lambda_2 \mathcal{L}_{T_{v \rightarrow q}},
\end{equation}
where $\lambda_1$ and $\lambda_2$ are the weight hyperparameters that regulate the loss of the feature transformer.

Finally, the training loss for IOG-VQA is as follows:
\begin{equation}
\mathcal{L}_{IOG} = \mathcal{L}_{GAN} + \alpha_1 \mathcal{L}_{WCE} + \alpha_2 \mathcal{L}_{distill},
\end{equation}
where $\alpha_1$ and $\alpha_2$ are loss hyperparameters.

\begin{table*}[t]
\centering
\caption{Experimental results on VQA-CP2 and VQA-CP1 test set. The \textbf{best} and \underline{second-best} results are styled in this way in the column. Compared with the same baseline architecture UpDn, our model IOG-VQA shows a relatively excellent performance.}
\label{table:2}
\begin{tabular}{lcccccccccc}
\hline
\multirow{2}{*}{\textbf{Models}} & \multirow{2}{*}{\textbf{Base}} & \multicolumn{4}{c}{\textbf{VQA-CP v2 test}} & \multirow{2}{*}{}  & \multicolumn{4}{c}{\textbf{VQA-CP v1 test}} \\
\cline{3-6}\cline{8-11}
& & \textbf{All} & \textbf{Y/N} & \textbf{Num.} & \textbf{Other} & \textbf{} & \textbf{All} & \textbf{Y/N} & \textbf{Num.} & \textbf{Other} \\
\hline
\textbf{SAN} \cite{34}  & - & 24.96 & 38.35 & 11.14 & 21.74 & & 32.50 & 36.86 & 12.47 & 36.22 \\
\textbf{GVQA} \cite{11} & - & 31.30 & 67.99 & 13.68 & 22.14 & & 39.23 & 64.72 & 11.87 & 24.86 \\
\textbf{S-MRL} \cite{6} & - & 38.46 & 42.85 & 12.81 & 43.20 & & 36.38 & 42.72 & 12.59 & 40.35 \\
\textbf{UpDn} \cite{14} & - & 39.94 & 42.46 & 11.93 & 45.09 & & 36.38 & 42.72 & 42.14 & 40.35 \\
\hline
\multicolumn{10}{l}{\textit{Methods based on modifying language modules follow:}} \\
\textbf{DLR} \cite{48} & UpDn & 48.87 & 70.99 & 18.72 & 45.57 & & 42.45 & 60.83 & 18.86 & 33.39 \\
\hline
\multicolumn{10}{l}{\textit{Methods based on strengthening visual attention follow:}} \\
\textbf{HINT} \cite{39} & UpDn & 46.73 & 67.27 & 10.61 & 45.88 & & 48.21 & 62.19 & 19.37 & 36.13 \\
\textbf{SCR} \cite{40} & UpDn & 49.45 & 72.36 & 10.93 & 48.02 & & - & - & - & - \\
\hline
\multicolumn{10}{l}{\textit{Methods based on balancing training data follow:}}  \\
\textbf{CVL} \cite{5} & UpDn & 42.12 & 45.72 & 12.45 & 48.34 & & 44.84 & 58.37 & 11.59 & 44.81 \\
\textbf{RandImg} \cite{25} & UpDn & 55.37 & 83.89 & 41.60 & 44.20 & & 57.24 & 76.53 & 33.87 & 46.57 \\
\hline
\multicolumn{10}{l}{\textit{Methods based on ensemble models follow:}}  \\
\textbf{AReg} \cite{9} & UpDn & 41.17 & 65.49 & 15.48 & 35.48 & & 43.43 & 74.16 & 12.44 & 25.32 \\
\textbf{RUBi} \cite{6} & UpDn & 44.23 & 67.05 & 17.48 & 39.61 & & 50.90 & 80.83 & 13.84 & 36.02 \\
\textbf{LMH} \cite{7} & UpDn & 52.45 & 69.81 & 33.44 & 45.54 & & 55.27 & 76.47 & 26.66 & 45.68 \\
\textbf{VGQE} \cite{31} & S-MRL & 50.11 & 66.35 & 27.08 & 46.77 & & - & - & - & - \\
\textbf{CF-VQA} \cite{10} & UpDn & 53.55 & \textbf{91.15} & 13.03 & 44.97 & & 57.03 & \textbf{89.02} & 17.08 & 41.47 \\
\textbf{GGE} \cite{8} & UpDn & 57.32 & 87.04 & 27.75 & 49.59 & & - & - & - & - \\
\textbf{Re-scaling} \cite{49} & UpDn & 47.09 & 68.42 & 21.71 & 42.88 & & 51.17 & 77.37 & 22.23 & 40.79 \\
\textbf{MDDC} \cite{32} & UpDn & 54.70 & 83.58 & 19.93 & 49.10 & & 58.31 & 82.58 & 24.67 & 46.98 \\
\textbf{GenB} \cite{33} & UpDn & 59.15 & 88.03 & 40.52 & 49.25 & & \textbf{62.74} & 86.18 & \underline{43.85} & \underline{47.03} \\
\textbf{TDS} \cite{TDS} & UpDn & 46.33 & 62.55 & 10.53 & 47.66 & & 45.55 & 58.22 & 13.82 & 45.70 \\
\textbf{PW-VQA} \cite{PW-VQA} & UpDn & 59.06 & 88.26 & \textbf{52.89} & 45.45 & & - & - & - & - \\
\textbf{CVIV+iter} \cite{50} & UpDn & 60.08 & 88.85 & 40.77 & \underline{50.30} & & - & - & - & - \\
\hline
\textbf{IOG-VQA (Ours)} & UpDn & \textbf{60.23} & \underline{89.14} & \underline{42.66} & \textbf{50.43} & & \underline{61.51} & \underline{86.32} & \textbf{46.51} & \textbf{47.64} \\
\hline
\multicolumn{10}{l}{\textit{Methods based on data augmentation follow:}}  \\
\textbf{KDDAug} \cite{28} & UpDn & 60.24 & 86.13 & 55.08 & 48.08 & & - & - & - & - \\
\textbf{DDG} \cite{DDG} & UpDn & 61.14 & 88.77 & 49.33 & 49.90 & & - & - & - & - \\
\textbf{Mutant} \cite{23} & UpDn & 61.72 & 88.90 & 49.68 & 50.78 & & - & - & - & - \\
\hline
\end{tabular}
\end{table*}

\subsection{Model Prediction and Probabilistic Results}

 Below are the Bias Model and Destination VQA Model predictions and the final output probabilities and results:
 The Bias Model learns the potential bias between the question and the visual features and outputs the corresponding probabilities. The Destination VQA Model combines visual and question features to make answer predictions and output probabilities, the formula for $P_b$ is similar to $P_d$ as follows:
 \begin{equation}
 P_b(A_i|Q_i, V_i) = \sigma(W_b \cdot f_b(Q_i, V_i)),
 \end{equation}
 \begin{equation}
 P_d(A_i|Q_i, V_i) = \sigma(W_d \cdot f_d(Q_i, V_i)),
 \end{equation}
 where $\sigma$ is the sigmoid function, $W_b$ and $W_d$ are the weights, $f_b$ and $f_d$ are the feature extraction functions of the Bias Model and Destination VQA Model, respectively, $P_b$ and $P_d$ are their output probabilities.
 The prediction probability of Destination VQA Model and Bias Model is weighted average to obtain the final output probability:
 \begin{equation}
 P(A_i|Q_i, V_i) = \beta P_d(A_i|Q_i, V_i) + (1 - \beta) P_b(A_i|Q_i, V_i),
 \end{equation}
 here $\beta$ is the weight parameter used to balance the output probabilities of the Destination VQA Model and Bias Model.
 Through the final output probability, the answer with the largest probability is selected as the final prediction result:
 \begin{equation}
 \hat{A}_i=\arg^{\max}_{A_i}P(A_i|Q_i,V_i),
 \end{equation}
 where $\hat{A}_i$ is the final predicted answer.

 Through the above modular process and its corresponding formulas and explanations, the proposed model can effectively reduce the bias and improve the overall performance of visual question-answering tasks by combining visual features and question features. The final output probability combines the prediction results of the Destination VQA Model and Bias Model to ensure the accuracy and robustness of the prediction.

\section{Experiments and Discussion}
\label{section4}

\subsection{Experiment Setting}
\subsubsection{Dataset Details}
We evaluate our model on the VQA-CP v2 and VQA-CP v1 datasets \cite{11}, which are designed for visual question answering under changing priors and have significantly different answer distributions between training and testing. VQA-CP v2 is created by rearranging and processing the VQA v2 dataset \cite{3}. 
TABLE \ref{table:1} shows the specific data amounts for the training and test sets of both VQA-CP v1 and VQA-CP v2, highlighting the differing answer distributions across question types.

In addition, the VQA-CP v1 and VQA-CP v2 datasets have emerged as the main benchmark datasets for studying bias in VQA models and are widely used in recent state-of-the-art methods. For example, recent state-of-the-art debiasing methods such as HINT \cite{39}, AReg \cite{9}, RUBi \cite{6}, LMH \cite{7}, CF-VQA \cite{10}, GenB \cite{33}, TDS \cite{TDS}, PW-VQA \cite{PW-VQA}, etc., all use VQA-CP v1 or VQA-CP v2 in their evaluations. These datasets provide a rigorous testing ground for evaluating debiasing strategies due to their intentionally skewed answer distributions. To further evaluate our method in a more comprehensive manner, we have also conducted experiments on the standard VQA v1 \cite{1} and VQA v2 \cite{3} datasets. The results of these additional experiments are reported in the paper, thereby strengthening our evaluation and demonstrating the broad applicability of our approach.

\subsubsection{Evaluation Settings}
We used the standard evaluation metrics from \cite{1} to assess our model. We also conducted experiments using the popular Bottom-up and Top-down Attention (UpDn) \cite{14} architecture, integrating it with our bias model and destination VQA model for comparison.

\begin{table*}[t]
\centering
\caption{EXPERIMENTAL RESULTS ON VQA V2 TEST / VAL AND VQA V1 TEST SET.}
\label{table:3}
\begin{tabular}{lcccccccccc}
\hline
\multirow{2}{*}{\textbf{Models}} & \multirow{2}{*}{\textbf{Base}} & \multicolumn{4}{c}{\textbf{VQA v2 test/val}} & \multirow{2}{*}{}  & \multicolumn{4}{c}{\textbf{VQA v1 test}} \\
\cline{3-6}\cline{8-11}
& & \textbf{All} & \textbf{Y/N} & \textbf{Num.} & \textbf{Other} & \textbf{} & \textbf{All} & \textbf{Y/N} & \textbf{Num.} & \textbf{Other} \\
\hline
\textbf{UpDn} \cite{14} & - & 63.48 & 81.18 & 42.14 & 55.66 & & - & - & - & - \\
\hline
\multicolumn{10}{l}{\textit{Methods based on ensemble models follow:}}  \\
\textbf{AReg} \cite{9} & UpDn & 62.75 & 79.84 & 42.35 & 55.16 & & 52.15 & 71.06 & 32.59 & 42.91 \\
\textbf{RUBi} \cite{6} & UpDn & 61.16 & - & - & - & & - & - & - & - \\
\textbf{LMH} \cite{7} & UpDn & 56.35 & 65.06 & 37.63 & 54.69 & & 48.78 & 61.25 & 30.18 & 42.74 \\
\textbf{VGQE} \cite{31} & S-MRL & 63.18 & - & - & - & & - & - & - & - \\
\textbf{CF-VQA} \cite{10} & UpDn & 63.54 & 82.51 & \textbf{43.96} & 54.30 & & 53.88 & 74.53 & 33.89 & 43.78 \\
\textbf{GGE} \cite{8} & UpDn & 59.11 & 73.27 & 39.99 & 54.39 & & - & - & - & - \\
\textbf{Re-scaling} \cite{49} & UpDn & 55.50 & 64.22 & 39.61 & 53.09 & & 48.12 & 60.74 & 31.15 & 41.82 \\
\textbf{MDDC} \cite{32} & UpDn & 63.33 & 81.64 & 42.56 & 54.88 & & 53.27 & 73.38 & 32.74 & 43.99 \\
\textbf{PW-VQA} \cite{PW-VQA} & UpDn & 62.63 & 81.80 & 43.90 & 53.01 & & - & - & - & - \\
\textbf{CVIV+iter} \cite{50} & UpDn & 61.42 & 79.34 & 39.85 & 53.50 & & - & - & - & - \\
\hline
\textbf{IOG-VQA (Ours)} & UpDn & \textbf{63.93} & \textbf{82.95} & 43.67 & \textbf{55.86} & & \textbf{54.64} & \textbf{75.25} & \textbf{34.64} & \textbf{44.28} \\
\hline
\end{tabular}
\end{table*}

\subsubsection{Implementation Details}
Our model, implemented in PyTorch, was fine-tuned for optimal performance using the Adam optimizer with a learning rate of 0.001, trained over 30 epochs with a batch size of 512. The architecture featured 1024 hidden units, alphabet embeddings with a dimension of 100, GloVe word embeddings with a dimension of 300 and a noise dimension is 2048. In the Object Interaction Self-Attention (OISA) module, the scaling factor was set to 64. Model parameters were regularly saved, and early stopping was used if performance did not improve over several epochs.

\subsection{Quantitative Results}
First, we compare our method IOG-VQA with the state-of-the-art or often-used comparison methods that also focus on bias reduction, as shown in TABLE \ref{table:2}. Past approaches are categorized in the following way: (1) Several baseline architectures such as SAN \cite{34}, GVQA \cite{11}, S-MRL \cite{6} and the one used in this article, UpDn \cite{14}, are listed in Section \ref{section1}. (2) The method of modifying the language module, the proposed language module suggests the separation of language concepts DLR \cite{48}. (3) Methods to enhance visual attention, which uses human vision \cite{37} or text \cite{38} explanation to enhance, including HINT \cite{39} and SCR \cite{40}. (4) The method of balancing training data changes the training distribution of unbiased training, including CVL \cite{5} and RandImg \cite{25}. (5) Ensemble model-based methods, including Areg \cite{9}, RUBi \cite{6}, LMH \cite{7}, VGQE \cite{31}, CF-VQA \cite{10}, GGE \cite{8}, Re-scaling \cite{49}, MDDC \cite{32}, GenB \cite{33}, CVIV+iter \cite{50} and our method IOG-VQA. We present the scores of these methods and their corresponding questions in TABLE \ref{table:2}. However, in the methods of balancing training data, to generate counterfactual training samples, some methods reduce or transform keywords and objects, such as CSS \cite{22}, or directly replace images CVL \cite{5}, RandImg \cite{25}. Data augmentation methods, such as KDDAug \cite{28}, DDG \cite{DDG}, and Mutant \cite{23}. It should be noted that while data augmentation methods focus on enhancing dataset diversity, our method primarily focuses on reducing dataset biases via GAN-based model debiasing. We did not compare these methods to balance the training data, because they change the training distribution, which is not the purpose of the VQA-CP dataset \cite{11} when it was proposed, which is to verify whether the VQA model is biased by the previously trained data.

TABLE \ref{table:2} presents the performance of the IOG-VQA model compared to several state-of-the-art methods on the VQA-CP v1 and VQA-CP v2 datasets.            The results indicate that IOG-VQA achieves superior performance across all categories when compared to existing methods, particularly in handling dataset biases, such as imbalanced distributions of answers.

\subsubsection{Performance on VQA-CP v2 Test Set}
On the VQA-CP v2 test set, IOG-VQA outperforms the baseline UpDn model by +20.29\% in the All category, achieving an accuracy of 60.23\% compared to UpDn’s 39.94\%. This significant improvement highlights the effectiveness of the Object Interaction Self-Attention (OISA) module and the Modified GAN-based debiasing in overcoming biases inherent in the dataset. In the Yes/No (Y/N) category, IOG-VQA achieves an accuracy of 89.14\%, which is an increase of +46.48\% over UpDn, which performs at 42.46\%. This demonstrates the model's ability to handle binary questions more robustly, likely due to the combination of object interaction understanding (via OISA) and debiasing mechanisms that reduce the model’s reliance on spurious correlations often found in biased datasets. For the Number (Num.) category, IOG-VQA is 1.89\% higher than the second-best method. This improvement is significant because the Number category often poses challenges in VQA tasks, especially when the model has to reason about quantities or counts from complex visual inputs. IOG-VQA’s ability to capture detailed visual relationships and reduce bias contributes to its enhanced performance on these types of questions. The most notable improvement is observed in the Other category, where IOG-VQA achieves 50.43\% accuracy, which is +5.34\% higher than UpDn’s 45.09\%. This improvement suggests that IOG-VQA is particularly effective at handling more complex and less frequently seen question types that require both nuanced image understanding and reasoning across objects. The OISA module, by modeling the interactions between objects in the image, likely plays a crucial role in this gain.

\begin{table*}[t]
\centering
\caption{Ablation of different training losses for our bias model. This table includes Weighted Cross-Entropy VQA Loss (WCE), Modified GAN Loss (GAN), and Kullback-Leibler Divergence Distillation Loss (Distill). The baseline combines WCE and the UpDn architecture. Using WCE alone significantly improves performance on VQA-CP v2 and VQA-CP v1. Adding GAN and Distill Losses further enhances all metrics, with the best results achieved when combining all three losses, indicating optimal bias perception and debiasing performance.}
\label{table:4}
\begin{tabular}{lcccccccccc}
\hline
\multirow{2}{*}{\textbf{Training Loss}} & \multirow{2}{*}{\textbf{Bias Model}} & \multicolumn{4}{c}{\textbf{VQA-CP v2 test}} & \multirow{2}{*}{}  & \multicolumn{4}{c}{\textbf{VQA-CP v1 test}} \\
\cline{3-6}\cline{8-11}
& & \textbf{All} & \textbf{Y/N} & \textbf{Num.} & \textbf{Other} & \textbf{} & \textbf{All} & \textbf{Y/N} & \textbf{Num.} & \textbf{Other} \\
\hline
\textbf{WCE}  & UpDn & 38.84 & 41.48 & 12.53 & 45.58 & & 38.80 & 40.44 & 12.38 & 42.89 \\
\hline
\textbf{WCE} & Ours & 56.41 & 88.38 & 19.44 & 49.16 & & 55.82 & 81.16 & 20.47 & 45.43 \\
\textbf{WCE + GAN} & Ours & 56.77 & 89.01 & 22.39 & 49.71 & & 57.72 & 82.39 & 35.21 & 46.28 \\
\textbf{WCE + Distill} & Ours & 57.24 & 88.79 & 24.83 & 49.69 & & 58.48 & 82.11 & 28.78 & 46.19 \\
\textbf{WCE + GAN + Distill} & Ours & \textbf{60.23} & \textbf{89.14} & \textbf{42.66} & \textbf{50.43} & & \textbf{61.51} & \textbf{86.32} & \textbf{46.51} & \textbf{47.64} \\
\hline
\end{tabular}
\end{table*}

\begin{table*}[t]
\centering
\caption{Ablation of ensemble debias loss on different bias models. This table includes three architectures: UpDn, LXMERT, and Ours Model. Each category compares its own combined debias loss and MDDC loss, as well as our loss. Additionally, it is compared with RUBi Loss and LMH Loss in our model architecture. Training with our loss improves metric scores across all three bias models.}
\label{table:5}
\begin{tabular}{ccccccccccc}
\hline
\multirow{2}{*}{\textbf{Bias Model}} & \multirow{2}{*}{\textbf{Ensemble Debias Loss}} & \multicolumn{4}{c}{\textbf{VQA-CP v2 test}} & \multirow{2}{*}{}  & \multicolumn{4}{c}{\textbf{VQA-CP v1 test}} \\
\cline{3-6}\cline{8-11}
& & \textbf{All} & \textbf{Y/N} & \textbf{Num.} & \textbf{Other} & \textbf{} & \textbf{All} & \textbf{Y/N} & \textbf{Num.} & \textbf{Other} \\
\hline
\multirow{3}{*}{\centering \textbf{UpDn}}
 & UpDn Loss & 39.94	& 42.46 & 11.93 & 45.09 & & 38.99 & 41.48 & 12.18 & 44.19 \\
& MDDC Loss & 54.70 & 83.58 & 19.93 & 49.10 & & 58.31 & 82.58	& 24.67	& 46.98 \\
& Our Loss & 52.47 & 88.20 & 30.09 & 40.38 & & 53.47 & 82.97 & 41.72 & 42.33 \\
\hline
\multirow{3}{*}{\centering \textbf{LXMERT}} & LXMERT Loss & 40.91 & 41.91 & 13.71 & 47.85 & & 36.38 & 42.72 & 12.59 & 40.35 \\
& MDDC Loss & 53.83 & 76.73 & 26.07 & 49.43 & & 59.44 & 83.21 & 26.78 & 46.73 \\
& Our Loss & 59.06 & 88.77 & 40.21 & 49.7 & & 59.69 & 83.88 & 40.75 & 46.99 \\
\hline
\multirow{4}{*}{\centering \textbf{Our Model}} & RUBi Loss & 30.77 & 72.78 & 12.15 & 13.84 & & 31.21 & 73.48 & 14.71 & 14.56 \\
& LMH Loss & 53.99 & 75.89 & 44.62 & 45.08 & & 54.10 & 76.10 & 45.17 & 46.98 \\
& MDDC Loss & 56.78 & 84.69 & 28.63 & 49.56 & & 59.49 & 83.71 & 28.53 & 47.11 \\
& Our Loss & \textbf{60.23} & \textbf{89.14} & \textbf{42.66} & \textbf{50.43} & & \textbf{61.51} & \textbf{86.32} & \textbf{46.51} & \textbf{47.64} \\
\hline
\end{tabular}
\end{table*}

\subsubsection{Performance on VQA-CP v1 Test Set}
On the VQA-CP v1 test set, IOG-VQA again shows strong performance improvements across all categories. In the All category, IOG-VQA achieves 61.51\%, which is +25.13\% better than UpDn’s 36.38\%. This indicates that the combination of OISA and GAN-based debiasing methods effectively enhances the model’s ability to handle the distributional shifts in the VQA-CP v1 dataset. For the Yes/No (Y/N) category, IOG-VQA achieves 86.32\%, showing a +43.6\% improvement over UpDn’s 42.72\%. This highlights that IOG-VQA is not only better at handling imbalanced datasets but also has a more generalized approach to answering binary questions, which are heavily influenced by dataset priors. IOG-VQA 2.66\% higher in the Num. category than the second-best method. The improvement in this category further reinforces the model’s ability to overcome the bias that frequently favors certain answer types over others, particularly when the model is faced with numerically-based questions. Lastly, IOG-VQA is 0.61\% higher in the Other category than the second-best method. This gain reflects the robustness of IOG-VQA in addressing more complex, open-ended questions that require the model to go beyond surface-level patterns and make more contextually grounded predictions.

\subsubsection{Performance on VQA v2 Test/Val and VQA v1 Test Sets}
In order to further demonstrate the generalization capability of our proposed IOG-VQA model, we conducted additional experiments on balanced (IID) datasets, namely VQA v2 and VQA v1. Table \ref{table:3} summarizes the performance of various methods on these datasets. For the VQA v2 test/validation set, our model built on the UpDn baseline achieves an overall accuracy of 63.93\%, with detailed scores of 82.95\% for the Yes/No questions, 43.67\% for the Number questions, and 55.86\% for the Other questions. On the VQA v1 test set, IOG-VQA attains an overall accuracy of 54.64\%, with category-wise accuracies of 75.25\% (Yes/No), 34.63\% (Number), and 44.28\% (Other). These results indicate that our IOG-VQA model not only effectively mitigates bias in the VQA-CP datasets but also maintains robust performance on balanced datasets. Compared with other ensemble-based approaches (e.g., AReg, RUBi, LMH, VGQE, CF-VQA, GGE, Re-scaling, MDDC, PW-VQA, and CVIV+iter), our method demonstrates competitive performance on VQA v2 and VQA v1. This superior performance is attributed to our dual mechanism that integrates object interaction self-attention with GAN-based debiasing, which effectively captures complex inter-object relationships while learning unbiased feature distributions. The comparable accuracies across both biased and balanced datasets suggest that the debiasing strategies integrated in IOG-VQA do not compromise, and may even enhance, the model's ability to generalize across different data distributions.

\subsubsection{Conclusion}
The quantitative results presented in Table 1 demonstrate the effectiveness of the IOG-VQA model in improving accuracy across multiple question categories. Compared to the baseline UpDn model, IOG-VQA shows consistent improvements, particularly in categories that are more prone to biases such as Yes/No and Other. The integration of the Object Interaction Self-Attention (OISA) module and the Modified GAN-based debiasing mechanism significantly enhances the model’s performance, especially in handling complex and nuanced visual reasoning tasks. The observed improvements across the All, Y/N, Num., and Other categories underscore the importance of addressing both object interactions and dataset biases in advancing the state of VQA models.

\subsection{Ablation Studies}

\begin{table*}[!t]
\centering
\caption{Ablation of different modules in the bias model. The results show that removing either the question processing module (Visual + Answer) or the visual processing module (Question + Answer) significantly reduces performance compared to the full model. The baseline architecture UpDn also shows lower performance. These results demonstrate the importance of integrating all components for optimal performance.}
\label{table:6}
\begin{tabular}{lccccccccc}
\hline
\multirow{2}{*}{\textbf{Training Loss}} & \multicolumn{4}{c}{\textbf{VQA-CP v2 test}} & \multirow{2}{*}{}  & \multicolumn{4}{c}{\textbf{VQA-CP v1 test}} \\
\cline{2-5}\cline{7-10}
& \textbf{All} & \textbf{Y/N} & \textbf{Num.} & \textbf{Other} & \textbf{} & \textbf{All} & \textbf{Y/N} & \textbf{Num.} & \textbf{Other} \\
\hline
\textbf{UpDn} & 39.94 & 42.46 & 11.93 & 45.09 & & 36.38 & 42.72 & 42.14 & 40.35 \\
\hline
\textbf{Visual + Answer} & 42.47 & 44.58 & 13.69 & 47.21 & & 42.69 & 48.34 & 31.74 & 45.67 \\
\textbf{Question + Answer} & 46.67 & 78.53 & 21.77 & 49.28 & & 50.73 & 72.18 & 23.37 & 46.67 \\
\textbf{Visual+Question+Answer} & \textbf{60.23} & \textbf{89.14} & \textbf{42.66} & \textbf{50.43} & & \textbf{61.51} & \textbf{86.32} & \textbf{46.51} & \textbf{47.64} \\
\hline
\end{tabular}
\end{table*}

\begin{figure*}[t]
    \centering
    \includegraphics[width=1\linewidth]{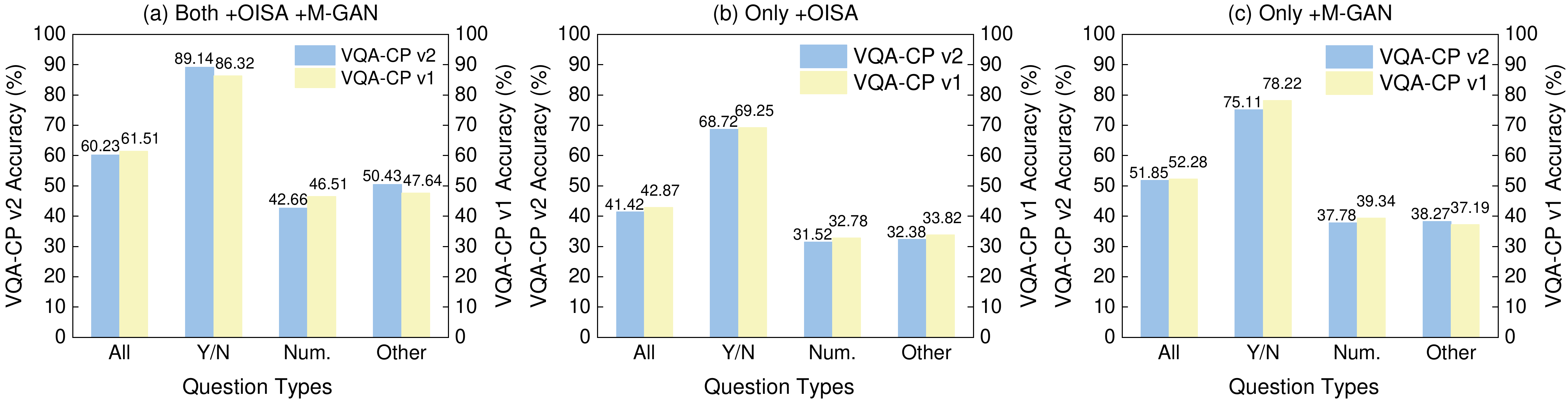}
    \caption{Comparison of the experimental results of module combination effect, where (a) represents the inclusion of both OISA module and Modified GAN module, (b) represents the inclusion of only OISA module but without Modified GAN module, (c) represents the inclusion of only Modified GAN module but without OISA module. Accuracy under different question types on VQA-CP v2 and VQA-CP v1, respectively.}
    \label{fig:merge}
\end{figure*}

We conducted a series of ablation studies to assess the impact of different components of our IOG-VQA model on VQA-CP v2 and VQA-CP v1 datasets. For comparison, all ablation study tables include experimental data from the baseline UpDn architecture. We analyzed the effects of various loss components on model performance and combined debiasing loss with bias models to demonstrate the superiority of our approach. It shows that the components we designed are effective and contribute to excellent performance.

\subsubsection{Analysis of the Train Loss}

We conducted ablation studies to evaluate the impact of different training losses on the performance of IOG-VQA, as shown in TABLE \ref{table:4}. The losses considered are: WCE (Weighted Cross-Entropy Loss), GAN (Modified GAN Loss), and Distill (Kullback-Leibler Divergence Distillation Loss). WCE Loss alone significantly improves performance on both VQA-CP v2 and VQA-CP v1, effectively addressing class imbalance by giving more weight to underrepresented classes. This shows the importance of balancing the loss contributions across different answer types. GAN Loss, added to WCE, further enhances the model's robustness by generating adversarial examples that help mitigate dataset biases. This improvement is particularly noticeable in categories like Number (Num.), where biases towards certain numerical answers are often present. Distill Loss, integrated with WCE and GAN, boosts generalization by transferring knowledge from a larger, more complex teacher model. This further refines the model’s predictions, leading to better performance across all categories. When combining all three losses, WCE + GAN + Distill yields the best overall performance, especially in Yes/No (Y/N) questions, where the model shows a nearly 50\% improvement over the UpDn baseline. Notably, WCE alone achieves near-optimal performance in the Other category, while the addition of GAN and Distill Loss significantly improves performance in the Number (Num.) category. These results demonstrate that the combination of these losses is crucial for handling dataset biases and enhancing the model’s ability to generalize across diverse question types.

\subsubsection{Analysis of the Ensemble Debias Loss}

In TABLE \ref{table:5}, we present results from experiments with various ensemble debias losses applied to bias models including UpDn, LXMERT \cite{21}, and our model. The debias losses evaluated are MDDC \cite{32}, RUBi \cite{6}, and LMH \cite{7}, representing popular methods for mitigating dataset bias in VQA. Our experiments show that integrating our proposed debias loss significantly enhances the performance of all models, both in accuracy and robustness. On the VQA-CP v2 dataset, our model improves UpDn by 12.53\%, reaching 52.47\% accuracy, compared to UpDn’s 39.94\%. When applied to LXMERT, our method results in an 18.15\% improvement, demonstrating its effectiveness even with models using advanced visual-linguistic strategies. Combining our loss with MDDC further boosts performance, especially in the Other and Number categories, where dataset bias is most prominent. This confirms that our debias model effectively addresses class imbalances and spurious correlations. Our approach also surpasses other debiasing methods, including RUBi and LMH, achieving the highest accuracy across all models. The improvements in both Yes/No and Number categories highlight the importance of a comprehensive debiasing strategy, including adversarial training (via GAN) and knowledge distillation. Overall, our results confirm the effectiveness of our ensemble debias loss, offering a robust solution to mitigate bias in VQA and advance the state-of-the-art.

\begin{figure*}[t]
    \centering
    \includegraphics[width=1\linewidth]{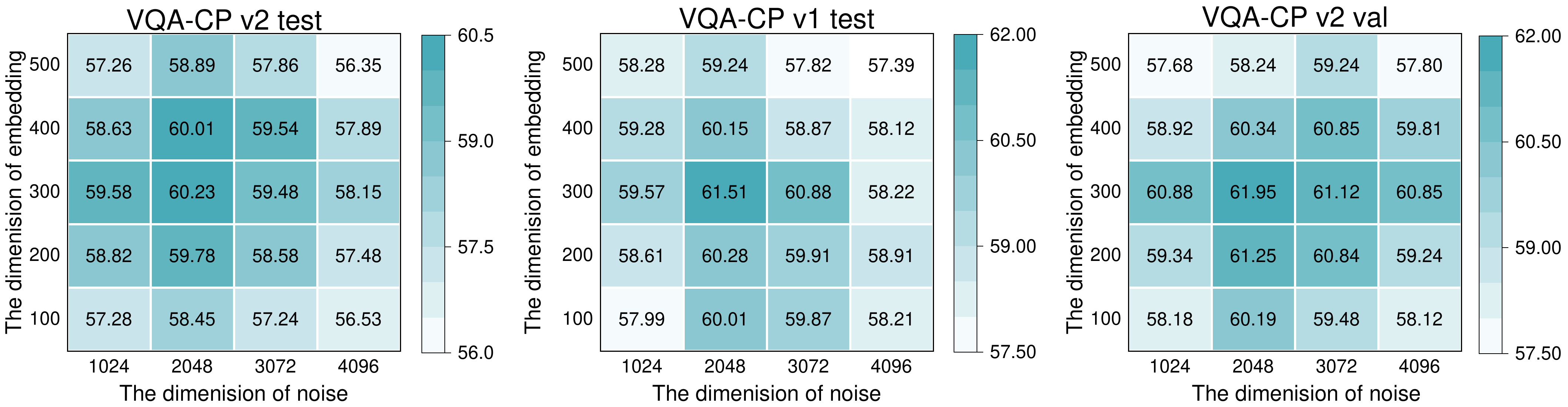}
    \caption{The influence of different embedding dimensions and noise dimensions on the model performance, respectively, on the test and validation sets of VQA-CP v2, and the test set of VQA-CP v1. }
    \label{fig:hot}
\end{figure*}

\begin{figure}[!t]
    \centering
    \includegraphics[width=0.9\linewidth]{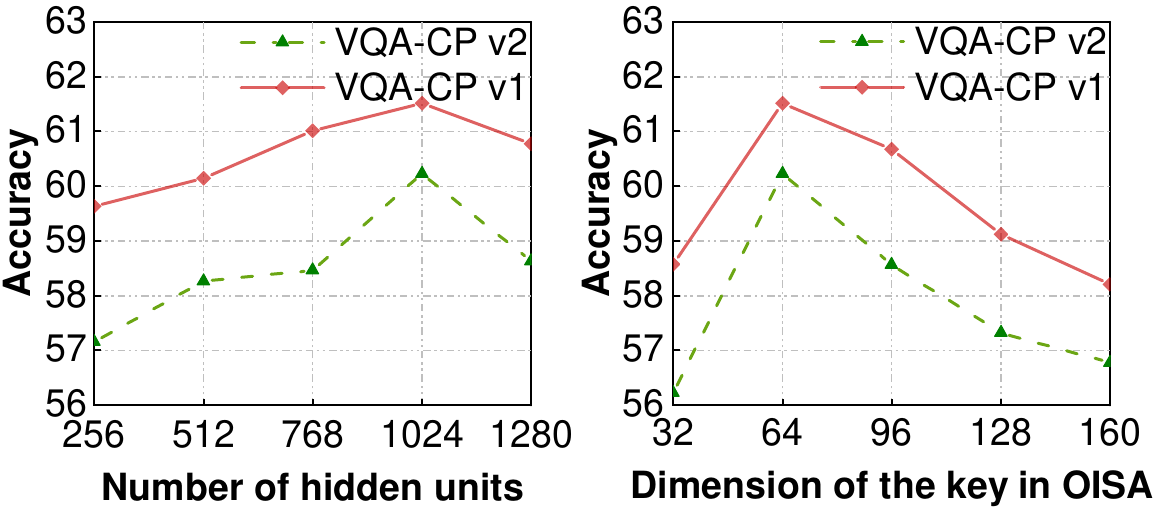}
    \caption{Hyperparameter tuning results. Accuracy peaks at 1024 hidden units and 64 key dimensions, optimizing performance on VQA-CP v2 and VQA-CP v1 datasets.}
    \label{fig:hyperparameter}
\end{figure}

\subsubsection{Ablation Analysis of Different Modules in the Bias Model}

To understand the impact of individual components, we conducted ablation experiments, shown in TABLE \ref{table:6}. These experiments removed key modules from the full model and assessed their effect on performance. The first row shows baseline results from the UpDn architecture, which serves as the foundation for our experiments. Using UpDn as the baseline, we created ablated versions, including the Visual + Answer and Question + Answer models. Visual + Answer Model: This model uses only visual input and answer labels, omitting the Question module. Without the Question module, the model cannot incorporate textual information, significantly reducing performance, as seen in TABLE \ref{table:6}. This highlights the importance of the Question module for grounding visual features in the context of the question. Question + Answer Model: This setup relies on the question and answer labels alone, without visual processing. The lack of visual input severely hinders performance, especially on tasks like object recognition, emphasizing the necessity of the Visual module for accurate VQA. Full Model (IOG-VQA): The full model, integrating both Visual and Question modules, consistently outperforms the ablated versions, achieving higher accuracy across all categories. These results confirm the superior performance of the full model, which leverages the interaction between visual and textual information for more accurate answers. The ablation experiments demonstrate that both the Visual and Question modules are crucial for effective VQA. Removing either significantly degrades performance, particularly for complex tasks. This reinforces the necessity of combining both visual and textual processing for improved VQA results.

\subsubsection{Comparison of the Experimental Results of Module Combination Effect}
In Fig. \ref{fig:merge}, the experiments are divided into three settings: (a) OISA combined with Modified GAN, (b) only OISA, and (c) only Modified GAN. The model is tested on four types of problems, "All", "Y/N", "Num.", and "Other" respectively, and the accuracy under each setting is recorded. Combining OISA and Modified GAN (Fig. \ref{fig:merge} (a)): shows the best accuracy on all question types, and especially shows significant debiasing effects on the biased "Other" question type. Using only OISA (Fig. \ref{fig:merge} (b)): The improvement is obvious in the "Other" type problem of understanding the interaction relationship between image objects, which proves that the OISA module can effectively improve the model's ability to capture complex visual relationships. Using only Modified GAN (Fig. \ref{fig:merge} (c)): performs well on "Y/N" and "Num." type problems, showing the effectiveness of the Modified GAN module in mitigating model bias. The combination of OISA and Modified GAN modules (Fig. \ref{fig:merge} (a)) achieves the highest accuracy on all kinds of problems, especially on the highly biased VQA-CP v2 dataset. This shows that the OISA module can enhance visual relationship understanding, while the Modified GAN module effectively reduces the dependence of the model on the bias of training data. Therefore, the combination of the two modules not only improves the visual reasoning ability of the model but also significantly improves the debiasing performance, which verifies the synergistic effect of OISA and Modified GAN modules.

\subsection{Sensitive Analysis}
\subsubsection{Sensitiveness of the Noise and Embedding Dimension}

As shown in Fig. \ref{fig:hot}, the goal of this experiment is to examine the impact of Embedding and Noise dimensions on model performance across different datasets to identify the optimal parameter combination. We conducted experiments on the test and validation sets of the VQA-CP v1 and VQA-CP v2 datasets, generating accurate heatmaps to visualize performance under various dimension configurations. The embedding dimension ranges from 100 to 500, and the noise dimension ranges from 1024 to 4096. For the VQA-CP v2 test and validation sets, accuracy improves with larger embedding and noise dimensions, particularly when the embedding is between 300 and 400, and the noise is around 2048. Further increases in dimensions show little to no performance improvement or slight decreases. For the VQA-CP v1 test set, high accuracy is achieved with embedding dimensions between 200 and 300 and noise dimensions between 2048 and 3072, indicating the model's reliance on specific dimension combinations. The results show that embedding and noise dimensions significantly affect performance. A moderate embedding dimension (e.g., 300) helps the model capture effective features, while a noise dimension around 2048 generates a stable, unbiased data distribution, enhancing model generalization. This configuration optimizes the model's debiasing ability and overall accuracy on the VQA-CP datasets.

\subsubsection{Hidden Units and OISA Key Dimension Sensitivity}
We conducted experiments to validate our hyperparameter settings, as shown in Fig. \ref{fig:hyperparameter}. We tested two hyperparameters: the number of hidden units in neural network layers and the key dimension in the Object Interaction Self-Attention (OISA) module on VQA-CP v2 and VQA-CP v1 datasets. The left figure shows that accuracy increases with the number of hidden units, peaking at 1024 before declining, indicating that 1024 hidden units optimize performance. The right figure shows that setting the key dimension to 64 achieves optimal accuracy, which decreases with higher dimensions, confirming 64 as the optimal key dimension for the OISA module.

\begin{table}[t]
\centering
\caption{Results of Ablation experiments under different combinations of Hyperparameter $\alpha_1$ and $\alpha_2$ on VQA-CP v2 test and VQA-CP v1 test set. (Only part of the values is shown.)}
\label{table:7}
\begin{tabular}{cccc}
\hline
$\alpha_1$ & $\alpha_2$ & \textbf{VQA-CP v2 test} & \textbf{VQA-CP v1 test} \\
\hline
\multirow{3}{*}{\centering 0.1}
& 0.1 & 58.52 & 59.08 \\
& 0.3 & 58.83 & 59.47 \\
& 0.5 & 59.18 & 59.86 \\
\hline
\multirow{3}{*}{\centering 0.3}
& 0.1 & 58.89 & 59.59 \\
& 0.3 & 59.34 & 60.21 \\
& 0.5 & 59.54 & 60.03 \\
\hline
\multirow{3}{*}{\centering 0.5}
& 0.1 & 59.99 & 60.73 \\
& 0.3 & \textbf{60.23} & \textbf{61.51} \\
& 0.5 & 60.05 & 61.08 \\
\hline
\multirow{3}{*}{\centering 0.7}
& 0.1 & 59.69 & 60.21 \\
& 0.3 & 59.73 & 60.61 \\
& 0.5 & 59.53 & 60.27 \\
\hline
\multirow{3}{*}{\centering 0.9}
& 0.1 & 59.54 & 60.05 \\
& 0.3 & 59.81 & 60.53 \\
& 0.5 & 59.23 & 60.25 \\
\hline
\end{tabular}
\end{table}

\subsubsection{Hyperparameter Sensitivity Analysis}
we have performed extensive grid search experiments on the VQA-CP v2 and VQA-CP v1 datasets to determine the optimal values for the hyperparameters $\beta$, $\alpha_1$, and $\alpha_2$, and presents the experimental results in Table \ref{table:7} and Table \ref{table:8}. $\beta$ (weighting the Destination VQA Model vs. the Bias Model outputs): We evaluated $\beta$ in the range [0.5, 0.9] with a step size of 0.1. Our experiments show that a value of $\beta$ = 0.7 achieves the best balance; lower values tend to overcorrect bias (thus degrading performance on balanced question types such as Yes/No), while higher values do not sufficiently mitigate dataset biases. $\alpha_1$ and $\alpha_2$ (loss weighting for $L_{\text{WCE}}$ and $L_{\text{distill}}$): We systematically tested $\alpha_1$ in the range [0.1, 1.0] and $\alpha_2$ in the range [0.1, 0.5]. Our experiments indicate that setting $\alpha_1$ = 0.5 and $\alpha_2$ = 0.3 yields the best overall performance. These values provide an optimal trade-off between correctly amplifying the loss contribution for rare answers and ensuring effective knowledge distillation without overpowering the GAN-based debiasing objective. As shown in Table \ref{table:7} and Table \ref{table:8}, the optimal performance is achieved when $\beta$ = 0.7, $\alpha_1$ = 0.5, and $\alpha_2$ = 0.3, yielding an overall accuracy of 60.23\% on VQA-CP v2 and 61.51\% on VQA-CP v1. These experiments confirm that our chosen hyperparameter settings optimally balance the contributions of the GAN loss, weighted cross-entropy loss, and distillation loss, and also ensure that the outputs of the Destination VQA Model and the Bias Model are more balanced.

\begin{table}[t]
\centering
\caption{Ablation Study on Overall Accuracy for VQA-CP v2 and VQA-CP v1 Test Datasets with Hyperparameter $\beta$.}
\label{table:8}
\resizebox{\columnwidth}{!}{%
\begin{tabular}{lccccccccc}
\hline
\multirow{2}{*}{\textbf{$\beta$}} & \multicolumn{4}{c}{\textbf{VQA-CP v2 test}} & \multirow{2}{*}{} & \multicolumn{4}{c}{\textbf{VQA-CP v1 test}} \\
\cline{2-5}\cline{7-10}
& \textbf{All} & \textbf{Y/N} & \textbf{Num.} & \textbf{Other} & \textbf{} & \textbf{All} & \textbf{Y/N} & \textbf{Num.} & \textbf{Other} \\
\hline
0.5 & 59.81 & 86.54 & 41.58 & 49.25 & & 60.07 & 83.51 & 45.59 & 46.52 \\
0.6 & 60.05 & 89.04 & 42.35 & 50.02 & & 60.83 & 86.01 & 46.09 & 47.08 \\
0.7 & \textbf{60.23} & \textbf{89.14} & \textbf{42.66} & \textbf{50.43} & & \textbf{61.51} & \textbf{86.32} & \textbf{46.51} & \textbf{47.64} \\
0.8 & 60.12 & 89.01 & 42.41 & 50.12 & & 61.03 & 86.14 & 46.21 & 47.33 \\
0.9 & 59.02 & 88.13 & 41.07 & 48.98 & & 59.46 & 85.14 & 45.18 & 46.28 \\
\hline
\end{tabular}
}
\end{table}

\subsection{Qualitative Results}

\begin{figure*}[t]
    \centering
    \includegraphics[width=0.93\linewidth]{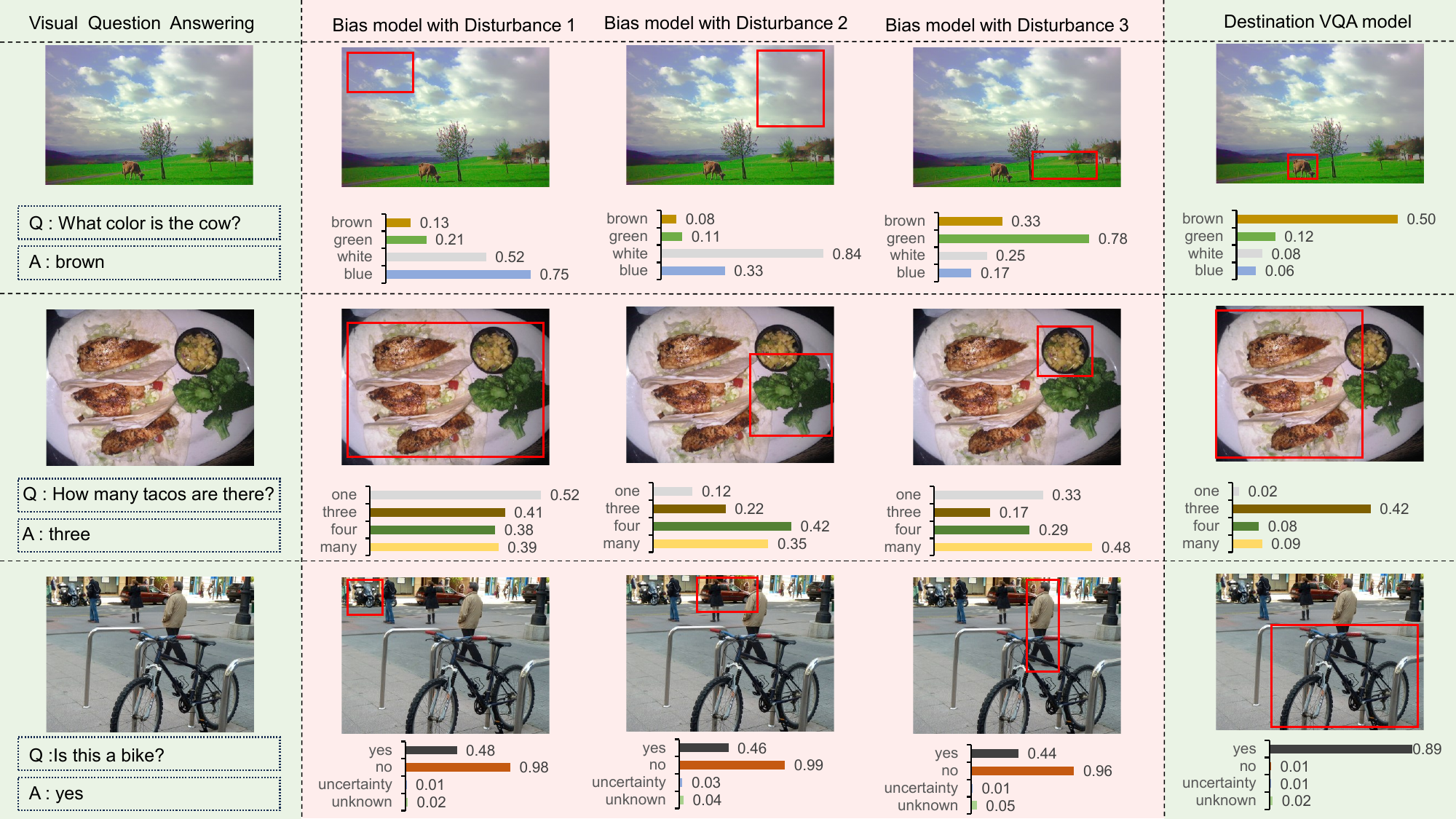}
    \caption{Our bias Model and Destination VQA Model are presented with predictions and scores through qualitative results. Green shows inputs and correct answers for three question types. Red highlights disturbance impacts with attention objects. Right green shows our model’s accurate predictions, demonstrating bias reduction and effective information processing.}
    \label{fig:case}
\end{figure*}

We visually present the initial Visual Question Answering data in Fig. \ref{fig:case}. As well as the specific results of the Bias model with the Disturbance and Destination VQA Model. The position of the green background on the left inputs the image question and the actual correct answer and three different types of questions are specially selected as examples, representing the three categories of Other, Num, and Y/N respectively. The position of the red background represents three times of Stochastic Disturbance to bias, where the red box represents the object of the picture. The object of attention is very different each time, so the possibility of leading to different results is also very different. The accuracy corresponding to the specific answer is shown at the bottom of each figure. The green background on the right is the prediction of the correct result, and the object detection shown by the red box highlights the ability of our model to reduce bias and combine all kinds of useful information to achieve excellent performance.

\section{Conclusion}
\label{section5}

In this paper, we presented the IOG-VQA model, which integrates Object Interaction Self-Attention and GAN-Based Debiasing to tackle biases in VQA tasks. Through extensive experiments on VQA-CP v1 and VQA-CP v2 datasets, we demonstrated that IOG-VQA outperforms existing methods, particularly in scenarios with imbalanced data distributions. Our approach effectively mitigates dataset bias while enhancing the model’s generalization ability by capturing complex inter-object relationships and fusing visual and textual cues.

In addition to these advantages, we acknowledge some limitations of our approach. For instance, the integration of a GAN-based debiasing framework with an object interaction module results in increased computational resource requirements and longer training times. Furthermore, while our method shows robust performance on standard benchmarks, there remains potential for further improvements in efficiency and scalability.

Future work will explore extending IOG-VQA to other visual reasoning tasks, reducing computational overhead, and further refining the debiasing mechanisms to achieve even greater robustness. We believe that addressing these limitations will pave the way for more efficient and widely applicable VQA systems.

\section*{Acknowledgments}
The authors sincerely thank the editors and anonymous reviewers for their constructive comments, which helped improve this paper. This work was supported in part by the National Natural Science Foundation of China (No. 62207011, 62377009, 62407013), the Natural Science Foundation of Hubei Province of China (No. 2025AFB653).


\bibliographystyle{Bibliography/IEEEtran}
\bibliography{Bibliography/Bibfile}

\begin{IEEEbiography}[{\includegraphics[width=1in, height=1.25in, clip, keepaspectratio]{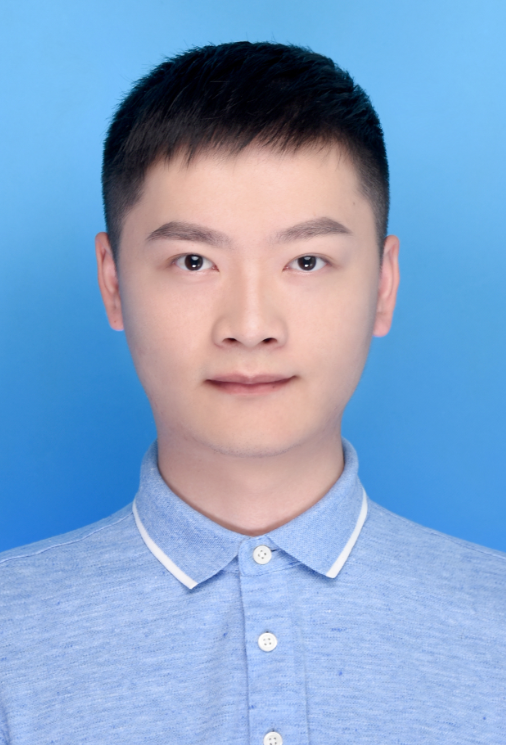}}] {Zhifei Li} received M.S. and Ph.D. degrees from the National Engineering Research Center for E-Learning at Central China Normal University in 2018 and 2021, respectively. Since October 2023, he has been an Associate Professor at the School of Computer Science, Hubei University. He has authored over thirty peer-reviewed papers in international journals and conferences, including TKDE, TNNLS, TMM, AAAI, and DASFAA, with three papers selected as ESI highly cited papers. Additionally, he frequently reviews for international journals and conferences, including TKDE, TNNLS, TKDD, and EMNLP. Since 2024, he has been a Communication Evaluation Expert for the NSFC. His current research interests include knowledge graphs and question answering.
\end{IEEEbiography}

\begin{IEEEbiography}[{\includegraphics[width=1in, height=1.25in, clip, keepaspectratio]{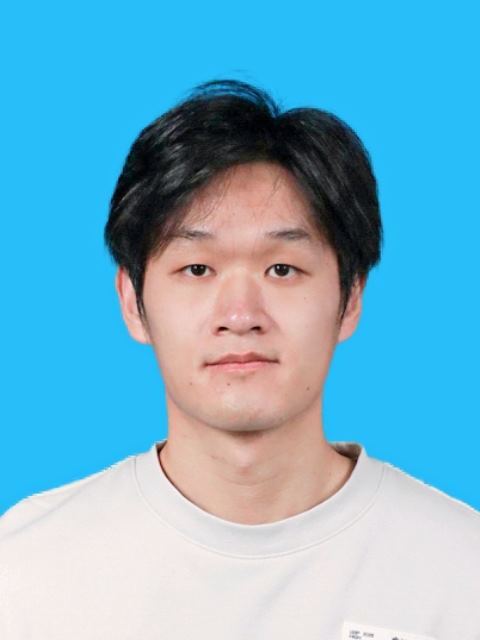}}] {Feng Qiu} received his B.S. degree from the Institute of Technology East China Jiao Tong University in 2022. He is currently pursuing a master's degree in the School of Computer Science at Hubei University. His research interests include visual question answering and knowledge graphs.
\end{IEEEbiography}

\begin{IEEEbiography}[{\includegraphics[width=1in, height=1.25in, clip, keepaspectratio]{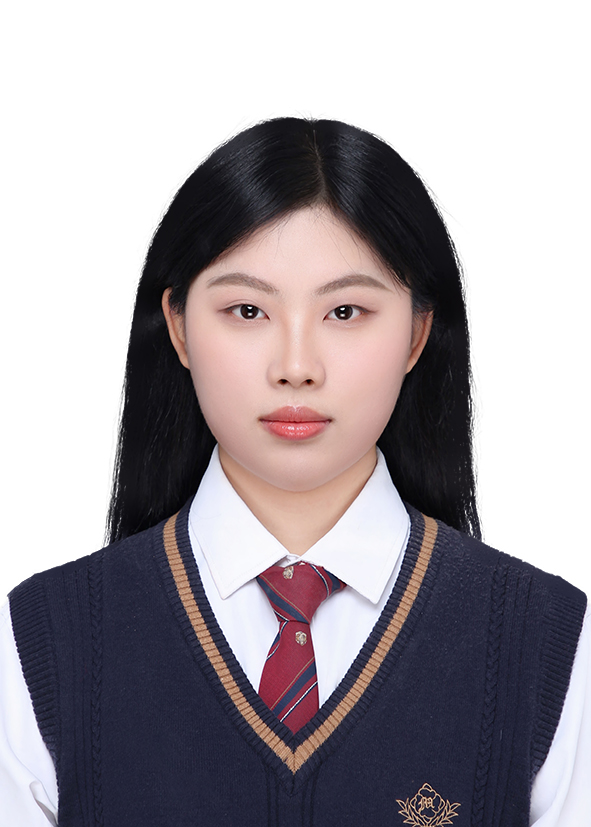}}] {Yiran Wang} received the B.S. degree from Xinyang Normal University in 2024. She is currently pursuing a master’s degree in the School of Computer Science at Hubei University. Her research interests include visual question answering, continual learning, and knowledge representation.
\end{IEEEbiography}

\begin{IEEEbiography}[{\includegraphics[width=1in, height=1.25in, clip, keepaspectratio]{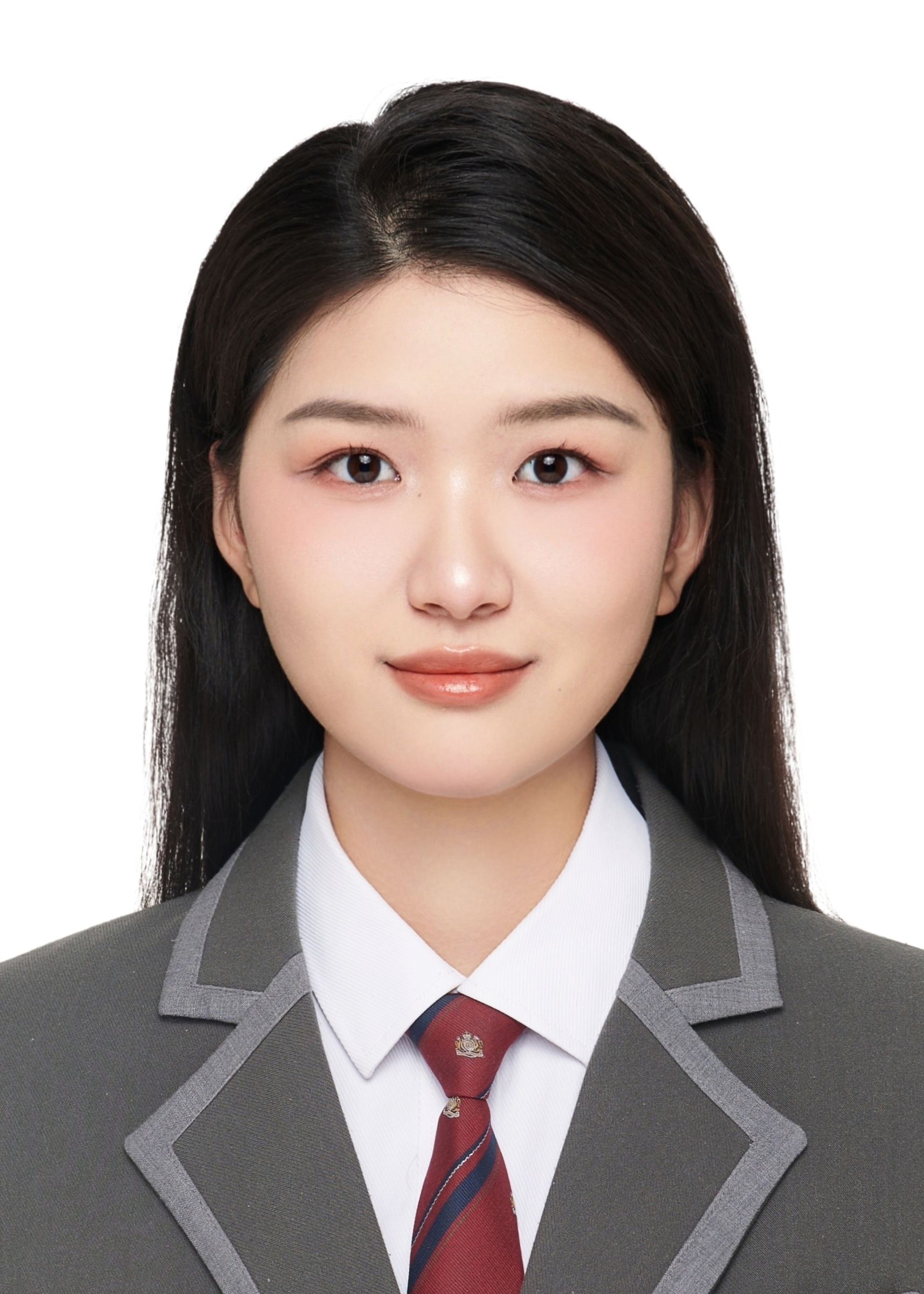}}] {Yujing Xia} received the B.S. degree from Hubei Engineering University in 2024. She is currently pursuing a master’s degree in the School of Computer Science at Hubei University. Her research interests include visual question answering, debiasing and dehallucination, and knowledge representation.
\end{IEEEbiography}

\begin{IEEEbiography}[{\includegraphics[width=1in, height=1.25in, clip, keepaspectratio]{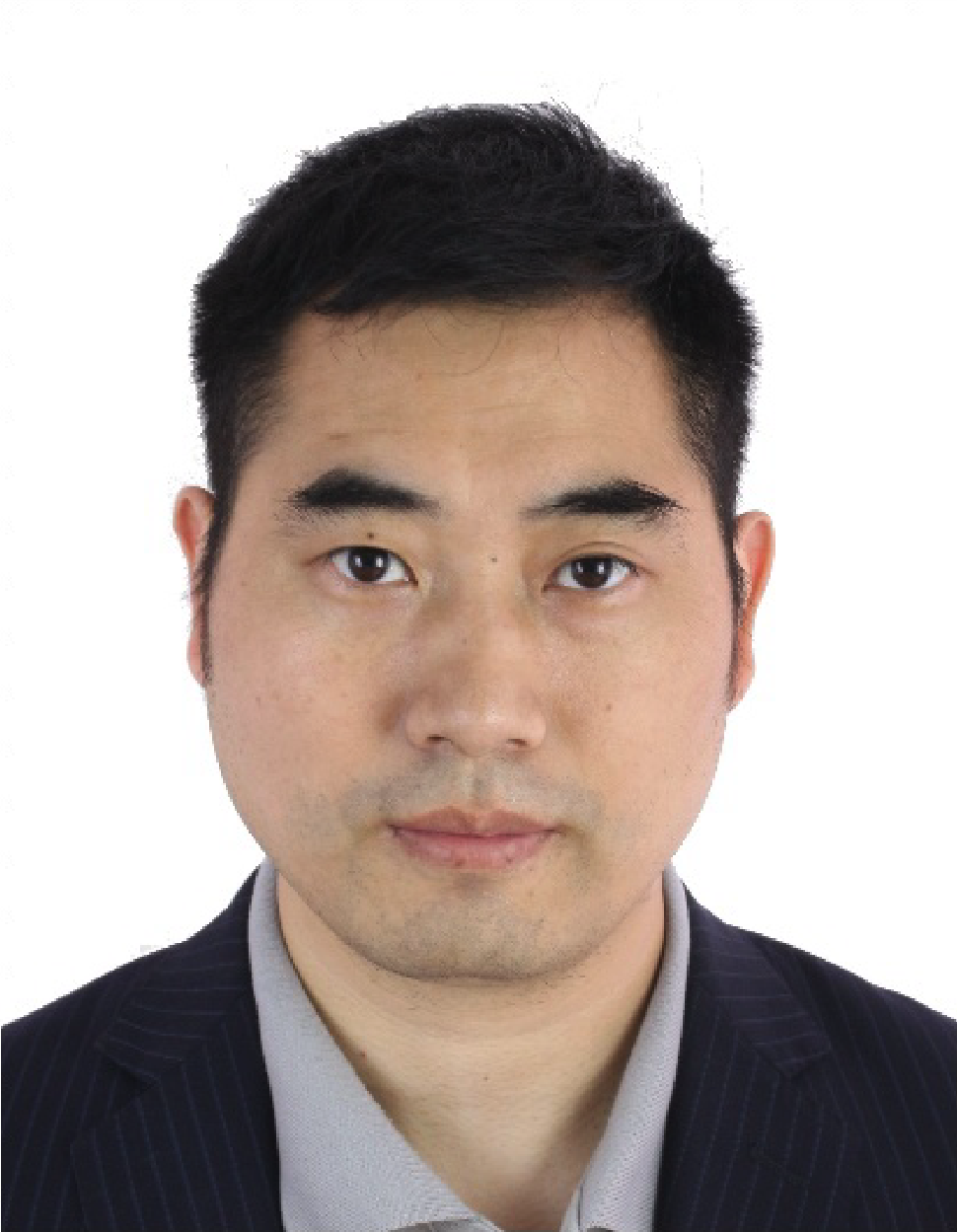}}] {Kui Xiao} received his Ph.D. degree in Wuhan University and is currently a professor in the School of Computer Science at Hubei University. He holds a senior member in the China Computer Federation (CCF).  His current research interests are artificial intelligence technology and educational data mining.
\end{IEEEbiography}

\begin{IEEEbiography}[{\includegraphics[width=1in, height=1.25in, clip, keepaspectratio]{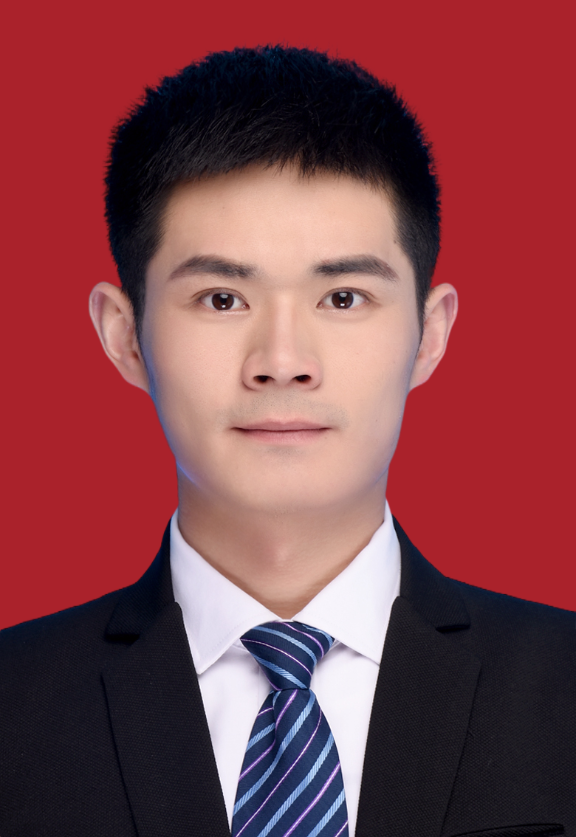}}]{Miao Zhang} received the Ph.D. degree from the National Engineering Research Center for E-Learning at Central China Normal University in 2023. He received the M.S. degree in computer technology from Central China Normal University, China in 2018. He is a lecturer in the School of Computer Science at Hubei University, China. His research interests include Q\&A systems and knowledge graphs.
\end{IEEEbiography}

\begin{IEEEbiography}[{\includegraphics[width=1in, height=1.25in, clip, keepaspectratio]{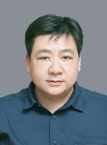}}]{Yan Zhang} received the Ph.D. degree from Beihang University. He is currently a professor in the School of Computer Science at Hubei University. His research interests include cyberspace security and software engineering, among which he has in-depth theoretical research and practical experience in source code security detection and verification.
\end{IEEEbiography}

\end{document}